\documentclass[letterpaper, 10 pt, conference]{ieeeconf}  

\IEEEoverridecommandlockouts                              

\usepackage{amsmath} 
\usepackage{graphicx}
\usepackage{cite}
\usepackage{hyperref}

\title{\LARGE \bf
CGiS-Net: Aggregating Colour, Geometry and Implicit Semantic Features for Indoor Place Recognition
}

\author{Yuhang Ming$^{1}$, Xingrui Yang$^{1}$, Guofeng Zhang$^{2}$ and Andrew Calway$^{1}$
\thanks{$^{1}$Yuhang Ming, Xingrui Yang and Andrew Calway are with the Visual Information Laboratory, Department of Computer Science,
        University of Bristol, Bristol, U.K.
        {\tt\small \{yuhang.ming, x.yang, andrew.calway\}@bristol.ac.uk}}%
\thanks{$^{2}$Guofeng Zhang is with the State Key Lab of CAD\&CG, Zhejiang University,
        Hangzhou, China.
        {\tt\small zhangguofeng@zju.edu.cn}}%
}

\begin{document}

\maketitle
\thispagestyle{empty}
\pagestyle{empty}

\begin{abstract}


 We  describe  a  novel  approach  to  indoor  place recognition   from RGB point clouds based on aggregating   low-level   colour   and   geometry features  with  high-level  implicit  semantic  features. It uses a 2-stage deep learning framework, in which the first stage is trained for the auxiliary task of semantic segmentation and the second stage uses features from layers in the first stage to generate discriminate descriptors for place recognition. The auxiliary task encourages the features to be semantically meaningful, hence aggregating the geometry and colour in the RGB point cloud data with implicit semantic information. We use  an  indoor place  recognition  dataset  derived  from  the  ScanNet  dataset for training and evaluation, with  a  test  set  comprising  3,608  point  clouds  generated  from 100  different  rooms.  Comparison  with  a  traditional  feature-based  method  and  four  state-of-the-art  deep  learning  methods demonstrate that our approach significantly outperforms all five methods, achieving, for example, a top-3 average recall rate of 75\%  compared  with  41\%  for  the  closest  rival  method. Our code is available at: \url{https://github.com/YuhangMing/Semantic-Indoor-Place-Recognition}
 
\end{abstract}

\section{INTRODUCTION}

Visual place recognition is a key capability to enable autonomous robots to operate in large-scale environments. It is an important research area in both robotics and computer vision and is frequently mentioned together with global localisation, serving as the first step prior to fine-grained pose estimation. It typically involves the generation of a global descriptor based on local features, followed by matching with those in a database of place-tagged descriptors. 

Previous work has focused primarily on place recognition for outdoor environments using RGB image data. This includes a large number of pre-deep learning methods and early convolutional neural network (CNN) approaches \cite{Survey}, as well as more recent end-to-end deep learning approaches inspired by NetVLAD \cite{NetVLAD}, such as \cite{Sarlin2019CVPR, Yu2020TNNLS, PatchNetVLAD}. More recently, PointNetVLAD \cite{PointNetVLAD} and its successors \cite{LPDNet, DH3D, MinkLoc3D} use 3-D point clouds as inputs and achieve very high average recall rates in outdoor environments. As a result, many large-scale outdoor place recognition datasets have been released with a focus on appearance and view-point differences \cite{247Dataset, CrossSeasonDataset, OxfordRobotCar}.

\begin{figure*}[t]
  \centering
  \includegraphics[width=0.98\textwidth]{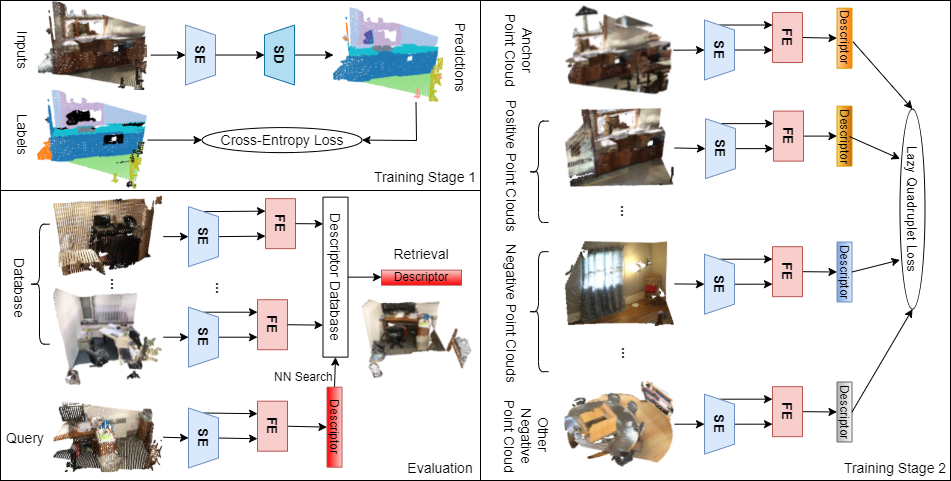}
  \vspace{-1ex}
  \caption{Overview of the CGiS approach, where SE denotes the semantic encoder, SD the semantic decoder and FE the feature embedding models. The models are trained in 2 stages: in stage 1 (top-left), the SE/SD models are trained for semantic segmentation using labelled RGB point cloud data and cross-entropy loss; in stage 2 (right), the FE model is trained to generate descriptors that discriminate between different places using levels from the pre-trained SE model and lazy quadruplet loss. For place recognition (bottom-left), descriptors generated from query point clouds are matched with those from a database of descriptors representing point clouds captured in different places using nearest neighbour (NN) search.}
  \label{fig::overview}
  \vspace{-3ex}
\end{figure*}


Comparatively less attention has been paid to place recognition in indoor environments. In many respects, the problem and challenges are similar, especially in terms of dealing with viewpoint and illumination changes, and the above approaches can be used. However, place recognition in indoor settings can often present different challenges that are not adequately addressed by these methods, resulting in poor performance, as the results in Table \ref{tab::compare} demonstrate. For example, it is often the case that query data only corresponds to a small part of a scene due to the close proximity of the sensor in, for example, a room environment, which contrasts with the wide vistas usually captured in outdoor applications. This limits the amount of information available for matching. In addition, indoor locations often have very similar appearances and structures, making discrimination especially difficult when using only RGB or point cloud data. Examples can be found in Fig. \ref{fig::layers} (third row) and Fig. \ref{fig::semantics} (last row).

In contrast, semantic information often provides greater discrimination when appearances and structures are ambiguous. For example, the entities “table” and “counter” may be structurally and sometimes visually similar, but are semantically different, often indicated by location context. We, therefore, hypothesise that using semantic features alongside low-level appearance and structural features will likely improve place recognition performance within indoor settings.

To investigate this we have developed a new approach to indoor place recognition that combines both colour and geometry features alongside high-level implicit semantic features. It is illustrated in Fig. \ref{fig::overview}. Inspired by the approach in \cite{Schonberger2018CVPR} designed for outdoor settings, we use an auxiliary semantic segmentation task to train a semantic encoder-decoder network, the features from different layers of which are then used to generate a global descriptor for place recognition via feature embedding. The auxiliary task encourages the network to learn semantically meaningful features, hence building semantic information alongside appearance and structure within the place descriptor. We use a 2-stage process to train the semantic encoder-decoder and feature embedding separately. 


There are no large-scale indoor place recognition datasets that support both images and point clouds. Hence we introduce a new one created from the ScanNet dataset \cite{ScanNet} for training and testing. It consists of 35,102 training point clouds generated from 565 different rooms, 9,693 validation point clouds from 142 rooms and 3,608 test point clouds from 100 rooms. Among the latter, 236 form the retrieval database and the remaining 3,372 make up the final test set. 
We present results that compare CGiS-Net with a hand-crafted feature solution and four deep learning approaches \cite{NetVLAD,PointNetVLAD,MinkLoc3D,DH3D} and show that it outperforms all 5 methods. 


To summarise, our contributions in this paper are three-fold: 1) we proposed a place recognition network that aggregates colour, geometry and implicit semantic features; 2) we derived an indoor place recognition dataset that supports both images and point clouds as inputs; 3) Both quantitative and qualitative experiments demonstrate that our CGiS-Net excels other popular place recognition networks.

\section{Related Work}

\subsection{Indoor place recognition}
Place recognition is commonly formulated as a retrieval problem and many works on indoor place recognition adopt the same formulation. 
\cite{Sizikova2016ECCVW} uses a Siamese network to simultaneously compute the features and descriptors from a pair of RGB-D frames.
Similar to RGB-D inputs, \cite{FDSLAM} modifies DH3D \cite{DH3D} for indoor scenes by introducing colour to the original point cloud inputs.
Using additional line features, LCD \cite{LCD} inputs both RGB-D frames and line clusters into the recognition network. Thus, enforcing that the learned global descriptors maintain structural information. Also utilising structural features, SpoxelNet \cite{SpoxelNet} extracts features at different levels and introduces a quad-view integrator on Lidar point clouds to handle the occlusion in the indoor environments. Our CGiS-Net also uses features at different levels but we only use a single extraction network while \cite{SpoxelNet} uses two separate feature extraction networks.

\subsection{Semantic place recognition}
Most semantic place recognition methods use explicitly represented semantic information. \cite{Frampton2013BMVC, Finman2014ICRA, Ming2021IROS} construct graphs of known objects or semantic entities to perform efficient place recognition. 
Operating on per-pixel semantic labels, \cite{Budvytis2019BMVC} generates the global descriptor with a histogram and \cite{Garg2019JRR} uses local semantic tensors to represent input images.
More recently, \cite{Neubert2021RSS} proposes a vector semantic representation that further encodes the layout of the semantic entities in a given input image. On the other hand, semantic information can also be implicitly incorporated into the global descriptor. \cite{VLASE} constructs global descriptors using NetVLAD layers with local features being the semantic edges extracted from the input images. Furthermore, \cite{Schonberger2018CVPR} trains an auto-encoder on a semantic scene completion task and then uses the latent code in-between the encoder and the decoder to create the implicit semantic vocabulary for place recognition. The network architecture of CGiS-Net proposed in this work is indeed inspired by these two works.

\begin{figure*}[t]
  \centering
  \includegraphics[width=0.98\textwidth]{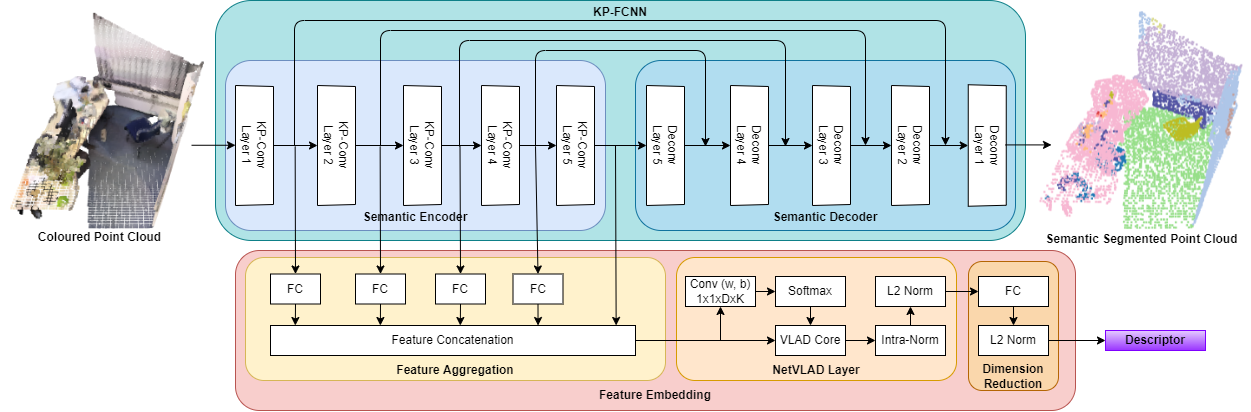}
  \vspace*{-1ex}
  \caption{The architecture of the proposed CGiS-Net.}
  \label{fig::architecture}
  \vspace*{-3ex}
\end{figure*}

\subsection{Indoor scene classification}
Indoor scene classification is similar to indoor place recognition but seeks to label room types rather than match data captured in the same room. Prior to deep learning, \cite{Orabona2007BMVC, Pronobis2010RAS} use handcraft features followed by SVM to perform classification. 
\cite{Song2019TIP} proposes to use separate CNNs to extract colour and depth features from RGB-D frames. Long short-term memory (LSTM) modules are followed to aggregate features learned over time. Also using separate networks to extract colour and depth features, \cite{Xiong2021TIP} further improves the classification performance by introducing a differentiable local feature selection module, achieving classification with single RGB-D frames. 
\cite{Du2019CVPR} proposes a network with one branch trained for the semantic segmentation task and the other branch for the classification task. In this way, the high-level semantic information is fully utilised in the classification task. Following this idea, \cite{Huang2020IROS} uses a 3-D CNN on the reconstruction of an entire room. This work is the closest one to our work but differs in the following aspects. First of all, \cite{Huang2020IROS} inputs an entire reconstruction of rooms built with a full sequence of RGB-D frames while our work only takes in a small point cloud generated from the views of single RGB-D frames. Secondly, given a query point cloud, we not only need to know which room the point cloud is captured in but also which part of the room it is captured. Finally, in addition to the high-level implicit semantic features, we also take advantage of the low-level colour and geometry features to boost the performance of indoor place recognition.


\section{Methodology}
\label{Sec::Method}

We follow the most popular place recognition formulation, casting the problem as a retrieval problem. Considering that the small appearance and structure changes in indoor scenes matter a lot, 
we choose to use RGB point clouds as the inputs to the network to fully utilise both colour and 3-D geometry information. 
Inspired by \cite{Schonberger2018CVPR}, we also propose to use implicit semantic features generated by a semantic encoder to achieve better indoor place recognition. 

\subsection{Network Architecture}
The architecture of the proposed CGiS-Net is illustrated in Fig. \ref{fig::architecture}. We choose to build the network on the state-of-the-art 3-D point cloud segmentation network, KP-FCNN \cite{KPConv} with deformed kernels, mainly because of its efficiency and robustness in handling input point clouds with various densities, thus providing more flexibility in the indoor place recognition task.
The proposed CGiS-Net consists of three main components, namely semantic encoder, semantic decoder and feature embedding models.

Following KP-FCNN, the semantic encoder comprises 5 KP-Conv layers with each one containing two convolutional blocks and the semantic decoder adopts the nearest upsampling. Skip connections are also introduced between corresponding encoder-decoder layers.
As it is reported in the KP-FCNN paper \cite{KPConv}, the lower KP-Conv layers in the semantic encoder tend to extract low-level geometry features such as corners and edges, and the latter KP-Conv layers usually focus more on complex and semantically meaningful features. We refer the readers to the original paper \cite{KPConv} for a detailed discussion on features extracted from different KP-Conv layers. 

We use features extracted from all the 5 KP-Conv layers in the semantic encoder to utilise all the low-level and high-level features. Before concatenating these multi-level features into a single feature map, fully connected (FC) layers are applied to stretch them into the same length. Then, the concatenated feature map is fed into a NetVLAD layer \cite{NetVLAD} to generate the place descriptor. To achieve more efficient retrieval operations, another FC layer is appended to the end of the NetVLAD layer for dimension reduction. 

\subsection{Multi-stage learning}
To ensure the features extracted by the latter KP-Conv layers in the encoder are semantically meaningful, 
we train the CGiS-Net in a 2-stage process.
Specifically, in the training stage 1, we train the semantic encoder and semantic decoder models on an auxiliary
semantic segmentation task in a standard supervised manner with the cross-entropy loss. 
We later validate that the features learned in-between the semantic encoder-decoder indeed contain semantic information in the Section \ref{sec::ablation}. And
because we don't use the explicit semantic segmentation results, we refer to the features used here as implicit semantic features. 


After the semantic encoder and semantic decoder are fully trained, 
we fix the weights of the semantic encoder and start training the feature embedding model in the training stage 2. Following PointNetVLAD \cite{PointNetVLAD}, metric learning with the lazy quadruplet loss is chosen to train the feature embedding model. 
The model inputs a tuple of an
anchor point cloud $P^{anc}$, $m$ positive point clouds $P^{pos} = \{P^{pos}_0, \dots, P^{pos}_{m-1}\}$, $n$ negative point clouds $P^{neg} = \{P^{neg}_0, \dots, P^{neg}_{n-1}\}$ and another negative point cloud which is negative to all the previous point clouds $P^{neg*}$, all of which are selected from the entire training dataset. 

When determining positive and negative point clouds, we use a criterion based on the distance between clouds, as in PointNetVLAD, and their scene ID. Specifically, given an anchor point cloud, a second point cloud is considered as a positive match if both point clouds are from the same scene and the distance between their centroids is less than a threshold $\tau^{pos}$.
If the two point clouds are from different scenes or the distance between them is larger than $\tau^{neg}$,
we say that the second point cloud is a negative match to the anchor point cloud. 
Note that we specify $\tau^{pos} < \tau^{neg}$ to maximise the difference between a negative pair. We will discuss the implementation and how to form the training tuples in detail later in the experiments section.

Once the tuples $\mathcal{T} = (P^{anc}, P^{pos}, P^{neg}, P^{neg*}) $ are generated, we can compute the lazy quadruplet loss as
\begin{equation}
\begin{split}
    \mathcal{L}_{LazyQuad}(\mathcal{T}) & = \underset{i,j}{\text{max}}([\alpha + \delta^{pos}_i-\delta^{neg}_j]_{+}) \\
    & + \underset{i,k}{\text{max}}([\beta + \delta^{pos}_i-\delta^{neg*}_k]_{+})
\end{split}
\vspace{-1ex}
\end{equation}
where $[\dots]_{+}$ denotes the hinge loss with constant margins $\alpha$ and $\beta$. $\delta^{pos}_i = d(P^{anc}, P^{pos}_i)$, $\delta^{neg}_j = d(P^{anc}, P^{neg}_j)$ and $\delta^{neg*}_k = d(P^{neg*}, P^{neg}_k)$ are the Euclidean distance between the point clouds.

\section{ScanNetPR Dataset}
We create the indoor place recognition dataset, ScanNetPR, from the annotated 3-D indoor reconstruction dataset, ScanNet \cite{ScanNet}. The ScanNet dataset contains 1,613 RGB-D scans of 807 different indoor scenes. It also provides rich semantic annotations with 20 semantic labels, making it perfect to test the proposed indoor place recognition network.
The whole dataset is divided into training, validation and test with 565, 142 and 100 scenes and 1,201, 312 and 100 scans respectively. Because the provided RGB-D frames are generated at the frame rate of 30 frames per second, the data is in fact very redundant and, depending on the movement of the RGB-D camera, there are tens or even hundreds of frames capturing the same place. Therefore, to make the data spatially sparser, we select keyframes from these scans based on the movement of the camera, both translationally and rotationally, resulting in 35,102 training keyframes, 9,693 validation keyframes and 3,608 test keyframes.

Then, the RGB point clouds are generated from these selected keyframes, forming the input of the proposed network. Rather than directly using the raw coloured point clouds back-projected from single RGB-D frames, we crop the coloured point clouds out of the complete reconstruction of the room using the viewing frustum of the given RGB-D frames. By doing so, we best alleviate the effect of the noisy depth measurements and the incomplete reconstruction of single views. 
We also store the RGB images and 3-D point clouds for each keyframe for comparison experiments.

\section{Experiments}

\begin{figure*}[t]
  \centering
  \includegraphics[width=0.98\textwidth]{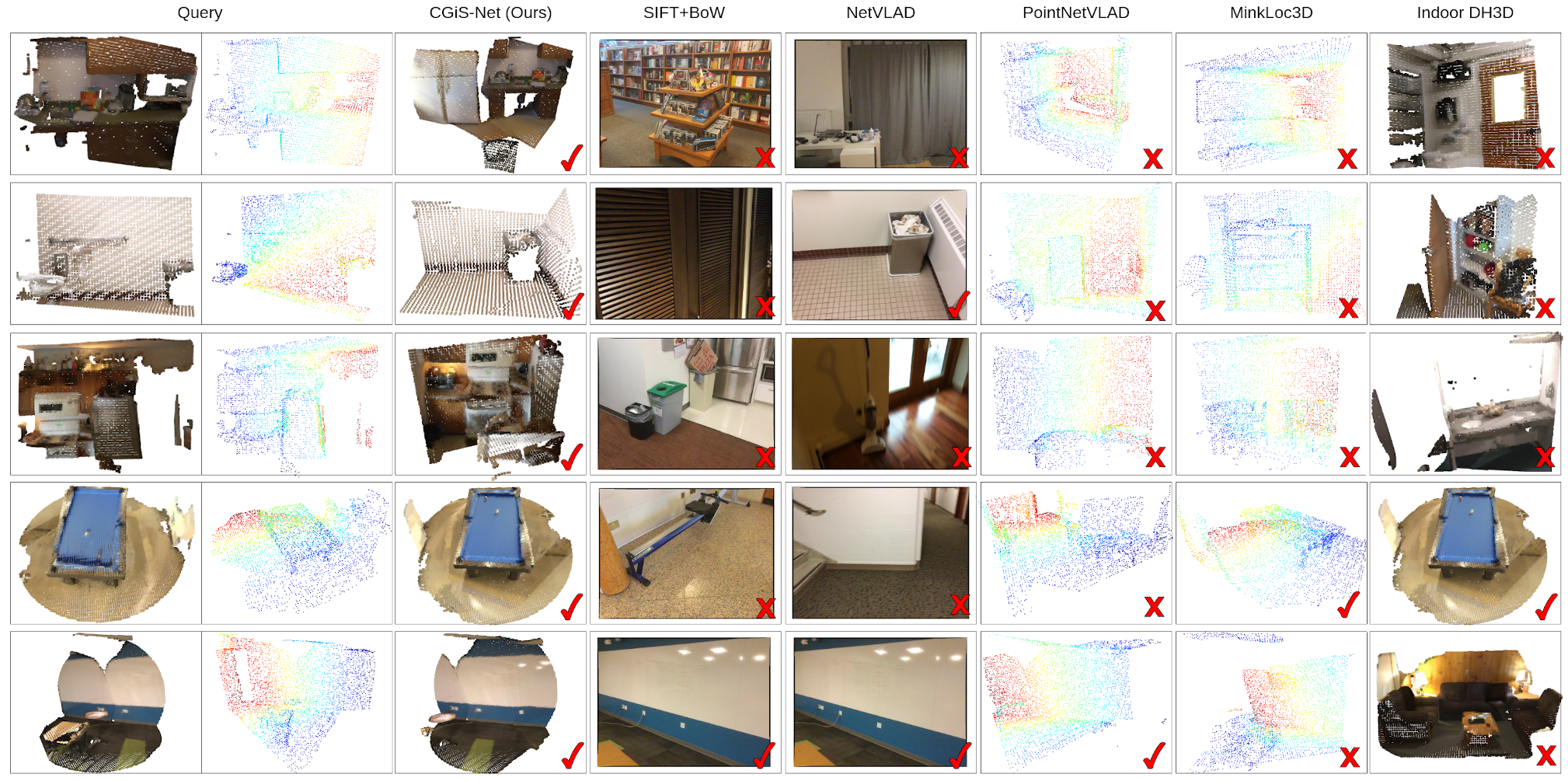}
  \caption{Examples of Top-$1$ retrievals with red checkmarks for succeeded ones and red crosses for the failed ones.
  The query entities are visualised in point clouds with and without colours and the retrieved database entities are visualised in the same form as their inputs, i.e. coloured point clouds for our method, images for SIFT \cite{SIFT} + BoW\cite{BoW} and NetVLAD \cite{NetVLAD}, and point clouds for PointNetVLAD \cite{PointNetVLAD}, MinkLoc3D \cite{MinkLoc3D} and indoor DH3D \cite{DH3D, FDSLAM}}
  \label{fig::results}
  \vspace*{-3ex}
\end{figure*}

\subsection{Training Procedure}
When training the semantic encoder and semantic decoder, we follow the SLAM segmentation setup in \cite{KPConv} and train the models with stochastic gradient descent (SGD) optimiser for 500 epochs. The general initial learning rate is set to 0.01 and the learning rate for deformed kernels is set to 0.001, both with learning rate decay applied. Momentum is also included in training with the value set to 0.98. Note that the whole ScanNet dataset is used in training the semantic encoder-decoder models to achieve the same semantic segmentation performance as reported in the original KPConv paper.

Then, to train the feature embedding model, the idea is to select as many negative point clouds when forming the training tuples. However, due to the memory limits on the hardware, we choose $m=2$ and $n=6$ in our implementation. Additionally, considering the size of indoor rooms, we set $\tau^{pos}=2m$ and $\tau^{neg}=4m$. The feature embedding module is trained with Adam optimiser for 60 epochs. The initial learning rate is set to 0.0001 and learning rate decay is also applied. To prevent overfitting, weight decay is also applied with the value set to 0.001. 
Following common choice on the hyperparameters of the NetVLAD layer\cite{PointNetVLAD, MinkLoc3D}, we set the
number of clusters $K=64$ and the output dimension $d_{out} = 256$.
The margin parameters in the lazy quadruplet loss are chosen to be $\alpha=0.5$ and $\beta=0.2$. 
With a single NVIDIA TITAN X, it takes around 24 hours to train the semantic encoder-decoder models and another 3 weeks to train the feature embedding model.

\subsection{Evaluation and Comparison}

To perform place recognition in the indoor environment which consists of 100 rooms from the test dataset, we first generate a database for later retrieval. Based on the distance between the point clouds, we store a new database point cloud if the new point cloud is either from a new scene or is at least 3 meters apart away from the previously stored database point clouds. In this way, we end up with 236 database point clouds and the rest 3,372 point clouds from the test dataset will be used as query point clouds.

Once we have the database point clouds, we obtain the final database descriptors by passing the database point clouds through the semantic encoder and the feature embedding models of CGiS-Net.
Then, given a query point cloud, the query descriptor is computed in the same way as the database ones.
Nearest neighbour search is performed between the query descriptor and database descriptors to retrieve $K$ nearest ones in the feature space. K-d tree is used for efficient search operations. On average it takes 0.095s to evaluate a query point cloud using an NVIDIA TITAN X.

We say the point cloud corresponding to the retrieved database descriptor is a correct match to the query point cloud if the two point clouds are from the same scene and the distance between them is less than 3 meters. Then the average recall rates of all query point clouds for Top-$K$ retrievals are computed and used as the main criterion for the evaluation.


In comparison, we first set a baseline performance with a pre-deep-learning method, which takes in RGB images, computes scale-invariant feature transform (SIFT) \cite{SIFT} and generates place descriptors with bag-of-words (BoW) \cite{BoW}. For deep-learning methods, we compare our CGiS-Net to the networks that have published their official implementations to avoid unfair comparison caused by re-implementation. 
We choose NetVLAD \cite{NetVLAD}, which uses RGB images as network input, PointNetVLAD \cite{PointNetVLAD} and MinkLoc3D \cite{MinkLoc3D}, which input point clouds, and an indoor-modification of DH3D \cite{DH3D, FDSLAM}, which takes in RGB point clouds. 

We re-trained these networks on our ScanNetPR dataset using the published training parameters, although we changed the loss function in NetVLAD to be the same as our CGiS-Net to make it a fair comparison and
left out the local feature detector and descriptor of DH3D as we are only interested in place recognition. Examples of queries and top-1 retrievals are shown in Fig. \ref{fig::results} and quantitative evaluations are provided in Table \ref{tab::compare}. The results show that CGiS-Net outperforms the other 5 methods to a large extent, demonstrating its effectiveness for indoor place recognition. Additionally, note that although the reported average recall rates of PointNetVLAD, MinkLoc3D and DH3D are very high for outdoor environments, their performance drops notably for our indoor dataset. Although care is needed when interrupting these results, especially since we did not optimise training parameters for indoor environments for these networks, we believe that the use of only geometry features taken from a limited number of input points is not sufficient to capture the detailed structural changes that discriminate between places, hence resulting in a significant reduction in recognition performance. This is supported by the fact that the indoor DH3D performs much better than PointNetVLAD and MinkLoc3D. We intend to investigate this further in future work.

\section{Ablation Study}
\label{sec::ablation}

\subsection{Colour features}
To prove that colour is crucial for indoor place recognition, we remove the RGB information from the input and re-train our CGiS-Net with only 3-D point clouds. Note that under this setup, we have to re-train not only the feature embedding model but also the semantic encoder-decoder models. 

The results are reported in the row ``CGiS-Net (w/o colour)'' of Table \ref{tab::compare}.
The performance of the CGiS-Net drops significantly without the additional colour in the input point clouds. We believe the reasons cause the degradation are two-fold. First of all, the lack of colour jeopardises the performance of the semantic segmentation, leading to inconsistent segmentation results. Hence, the final place recognition performance is also jeopardised. Secondly, the variety of the structural complexity and the structural similarities of indoor scenes are higher compared to those of outdoor scenes, making it not distinguishable enough to only use 3-D point clouds for indoor place recognition. 

\begin{table}[t]
\caption{Average Recall Rate}
\label{tab::compare}
\vspace*{-3ex}
\begin{center}
\begin{tabular}{|c||c|c|c|}
\hline
 & Recall @1 & Recall @2 & Recall @3 \\
\hline
\hline
SIFT \cite{SIFT} + BoW \cite{BoW} & 16.16\% & 21.17\% & 24.38\% \\
\hline
NetVLAD \cite{NetVLAD} & 21.77\% & 33.81\% & 41.49\% \\
\hline
PointNetVLAD \cite{PointNetVLAD} & 5.31\% & 7.50\% & 9.99\% \\
\hline
MinkLoc3D \cite{MinkLoc3D} & 3.32\% & 5.81\% & 8.27\% \\
\hline
Indoor DH3D \cite{DH3D,FDSLAM} & 16.10\% & 21.92\% & 25.30\% \\

\hline
\hline
CGiS-Net (default) & \textbf{61.12\%} & \textbf{70.23\%} & \textbf{75.06\%} \\
\hline
CGiS-Net (w/o colour) & 39.62\% & 50.92\% & 56.14\% \\
\hline
CGiS-Net (w/o geometry) & 40.07\% & 51.28\% & 58.96\% \\
\hline
CGiS-Net (w/o semantics) & 54.95\% & 64.49\% & 70.36\% \\
\hline
CGiS-Net (fixed density) & 31.58\% & 42.29\% & 48.93\% \\
\hline
\end{tabular}
\end{center}
\vspace*{-6ex}
\end{table}
 
\subsection{Geometry features}
\begin{figure}[t]
  \centering
  \includegraphics[width=0.47\textwidth]{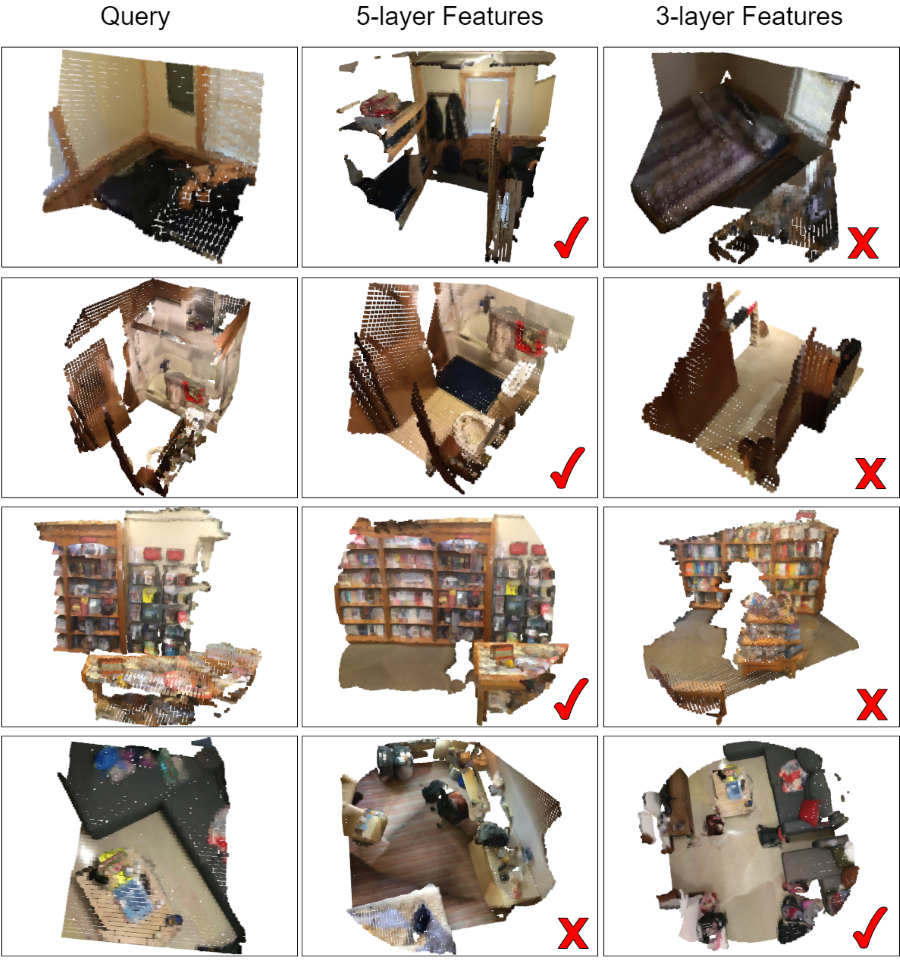}
  \caption{Examples of Top-$1$ retrievals with the CGiS-Net trained with local features from 5 KP-Conv layers and 3 KP-Conv layers.}
  \label{fig::layers}
  \vspace{-3ex}
\end{figure}

In the default setting, 
we concatenate features extracted from all the 5 KP-Conv layers of the semantic encoder. However, in this experiment, we focus on the features with semantic meanings and only concatenate the features from the last 3 KP-Conv layers.

Examples of the top-1 retrieved point clouds by the CGiS-Net trained with 5-layer features and 3-layer features are shown in Fig. \ref{fig::layers}. We observe that the network trained with 3-layer features tends to find point clouds that contain the same semantic entities like the ones in the query, such as ``bed'', ``door'' and ``bookshelf'' from the first three examples. However, only focusing on the semantic entities is not enough as the same semantic entities exist in different rooms. Utilising the additional low-level features is necessary to achieve better performance.
The quantitative results are provided in the row ``CGiS-Net (w/o geometry)'' of Table \ref{tab::compare}, which demonstrate that using the geometry features extracted from the first 2 KP-Conv layers indeed boosts the recognition performance.
However, we have to admit that the network using all 5-layer features sometimes gets lost in the tiny details in the scene due to the unbalanced feature size while focusing on the semantics, like ``sofa'' in the last example, can retrieve the correct database point cloud.



\subsection{Implicit semantic features}
\begin{figure}[t]
  \centering
  \includegraphics[width=0.5\textwidth]{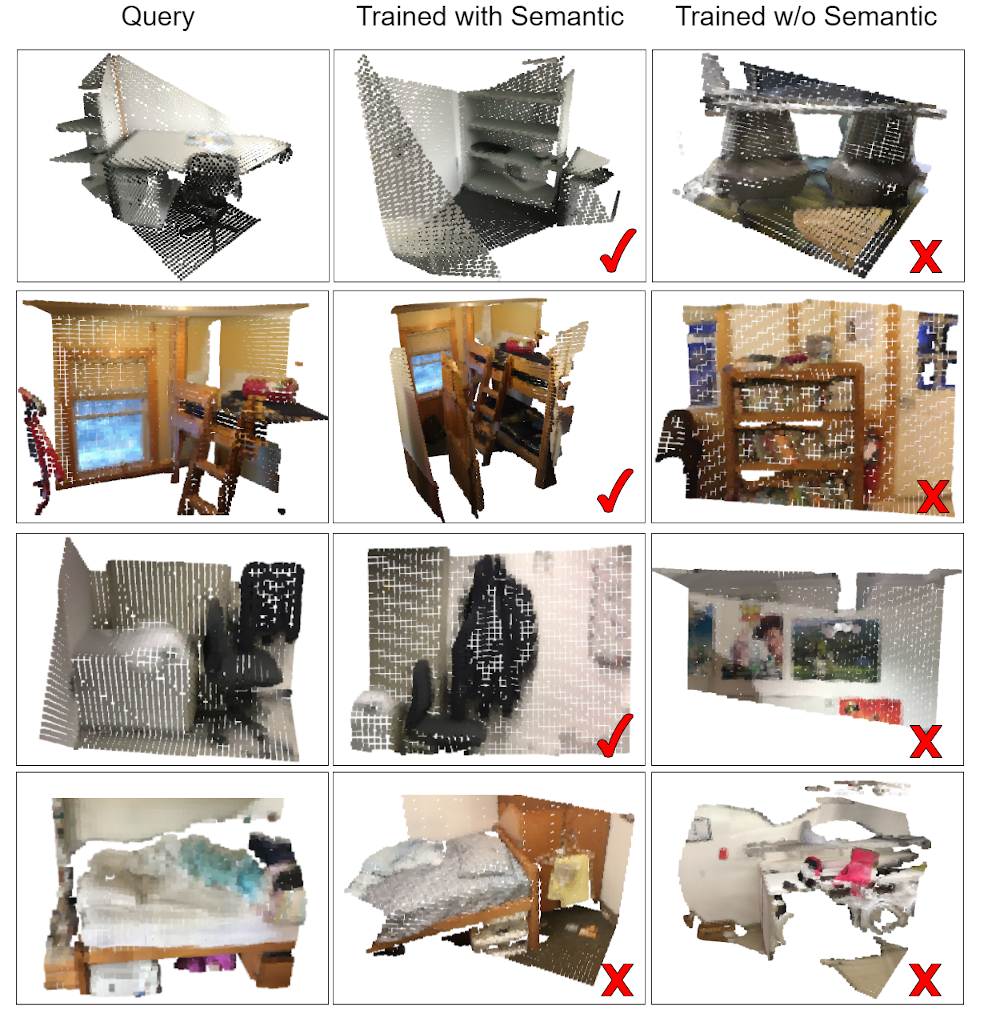}
  \caption{Examples of Top-$1$ retrievals with the CGiS-Net trained with semantic segmentation and without semantic segmentation.}
  \label{fig::semantics}
  \vspace{-3ex}
\end{figure}

To further investigate how much improvement in indoor place recognition performance is brought by the implicit semantic features learned in the semantic segmentation task, we re-design the architecture of the proposed network by removing the semantic decoder while keeping the encoder and the feature embedding models. 
Under this setup, multi-stage learning doesn't fit anymore. We re-train the new network
using only the second training stage
with the same lazy quadruplet loss as the original network and the Adam optimiser for 60 epochs. 

The quantitative evaluation results are reported in the row ``CGiS-Net (w/o semantics)'' of Table \ref{tab::compare}. 
Compared to the default model, we observe a roughly consistent 5\% drop in average recall rates, demonstrating the importance of the implicit semantic features in indoor place recognition. 
Additionally, we also provide examples of top-1 retrieved point clouds by the network trained with and without semantics, shown in Fig. \ref{fig::semantics}. The first 3 rows in the figure show the cases when the network trained without semantics failed while the originally proposed network, i.e. trained with semantics, succeeded. We can observe that without forcing the network to learn implicit semantic features, the retrieved point clouds tend to be similar to the query point clouds in terms of appearances and 3-D structures. However, as we already demonstrated in the previous sections, indoor scenes contain a lot of entities with similar appearances or structures but completely different semantic meanings. Therefore, implicit semantic features are indispensable for robust indoor place recognition. In the last row in Fig. \ref{fig::semantics},
although the models failed to retrieve the correct place, 
We can still observe that the network trained with implicit semantic features tries to find point clouds not only with similar colours and structures but also the same semantic object, \textit{i.e.} ``bed'', in the scene.

\subsection{Point cloud densities}
KP-FCNN \cite{KPConv} is good at handling point cloud inputs of various densities. However, PointNet \cite{PointNet} used in PointNetVLAD \cite{PointNetVLAD}, MinkLoc3D \cite{MinkLoc3D} and indoor DH3D \cite{DH3D, FDSLAM} can only take in point clouds with a fixed number of 4096 or 8192 points. To make it a fair comparison and also to prove that denser points lead to better recognition performance, we re-train CGiS-Net with the same input as the PointNetVLAD and the MinkLoc3D, i.e. point clouds with a fixed number of 4096 points and without colour. In this setup, the semantic encoder-decoder models also need to be re-trained.

The results of this training setup are shown in the row ``CGiS-Net (fixed density)'' of Table \ref{tab::compare}. Compared to the results from the model trained without colours, the recognition performance suffers a considerable drop, especially for the Top-$1$ average recall rate. The results prove that denser point clouds are preferred to better capture 3-D geometry features of indoor environments. On the other hand, although the performance is worse compared to the default training setup, it is still much better compared to the PointNetVLAD, MinkLoc3D and indoor DH3D. 

\section{Conclusions}

We have proposed CGiS-Net for indoor place recognition based on aggregating colour, geometry and implicit semantic features to learn global descriptors. Using an indoor place recognition dataset derived from the ScanNet dataset, we showed that performance exceeds a traditional feature-based method and four recently proposed place recognition networks. In future, we intend to investigate the use of attention modules to handle the unbalanced size of features from different KP-Conv layers and explore performance in greater detail using additional indoor and outdoor datasets. 


\begin{thebibliography}{99}


\bibitem{Survey} S. Lowry, N. Sunderhauf, P. Newman, J. J. Leonard, D. Cox, P. Corke and M. J. Milford, ``Visual Place Recognition: A Survey,'' in \textit{IEEE Transactions on Robotics}, vol. 32, no. 1, pp. 1-19, 2016.

\bibitem{NetVLAD} R. Arandjelović, P. Gronat, A. Torii, T. Pajdla and J. Sivic, ``NetVLAD: CNN Architecture for Weakly Supervised Place Recognition,'' in \textit{IEEE Transactions on Pattern Analysis and Machine Intelligence}, vol. 40, no. 6, pp. 1437-1451, 2018.


\bibitem{Sarlin2019CVPR} P. Sarlin, C. Cadena, R. Siegwart and M. Dymczyk, ``From Coarse to Fine: Robust Hierarchical Localization at Large Scale,'' in \textit{IEEE/CVF Conference on Computer Vision and Pattern Recognition}, pp. 12708-12717, 2019.


\bibitem{Yu2020TNNLS} J. Yu, C. Zhu, J. Zhang, Q. Huang and D. Tao, ``Spatial Pyramid-Enhanced NetVLAD with Weighted Triplet Loss for Place Recognition," in \textit{IEEE Transactions on Neural Networks and Learning Systems}, vol. 31, no. 2, pp. 661-674, 2020.

\bibitem{PatchNetVLAD} S. Hausler, S. Garg, M. Xu, M. Milford and T. Fischer, ``Patch-NetVLAD: Multi-Scale Fusion of Locally-Global Descriptors for Place Recognition,'' in \textit{IEEE/CVF Conference on Computer Vision and Pattern Recognition}, pp. 14141-14152, 2021.

\bibitem{PointNetVLAD} M. A. Uy and G. H. Lee, ``PointNetVLAD: Deep Point Cloud Based Retrieval for Large-Scale Place Recognition,'' in \textit{IEEE/CVF Conference on Computer Vision and Pattern Recognition}, pp. 4470-4479, 2018.


\bibitem{LPDNet} Z. Liu, S. Zhou, C. Suo, P. Yin, W. Chen, H. Wang, H. Li and Y. Liu, ``LPD-Net: 3D Point Cloud Learning for Large-Scale Place Recognition and Environment Analysis,'' in \textit{IEEE/CVF International Conference on Computer Vision}, pp. 2831-2840, 2019.

\bibitem{DH3D} J. Du, R. Wang and D. Cremers, ``DH3D: Deep Hierarchical 3D Descriptors for Robust Large-Scale 6DoF Relocalization,'' in \textit{European Conference on Computer Vision}, 2020.



\bibitem{MinkLoc3D} J. Komorowski, ``MinkLoc3D: Point Cloud Based Large-Scale Place Recognition,'' in \textit{IEEE Winter Conference on Applications of Computer Vision}, pp. 1789-1798, 2021.

\bibitem{247Dataset} A. Torii, R. Arandjelović, J. Sivic, M. Okutomi and T. Pajdla, ``24/7 Place Recognition by View Synthesis,'' in \textit{IEEE Conference on Computer Vision and Pattern Recognition}, pp. 1808-1817, 2015.

\bibitem{CrossSeasonDataset} M. Måns Larsson, E. Stenborg, L. Hammarstrand, M. Pollefeys, T. Sattler and F. Kahl, ``A Cross-Season Correspondence Dataset for Robust Semantic Segmentation,'' in \textit{IEEE/CVF Conference on Computer Vision and Pattern Recognition}, pp. 9524-9534, 2019.

\bibitem{OxfordRobotCar} D. Barnes, M. Gadd, P. Murcutt, P. Newman and I. Posner, ``The Oxford Radar RobotCar Dataset: A Radar Extension to the Oxford RobotCar Dataset,'' in \textit{IEEE International Conference on Robotics and Automation}, pp. 6433-6438, 2020.


\bibitem{Schonberger2018CVPR} J. L. Sch\"onberger, M. Pollefeys, A. Geiger and T. Sattler, ``Semantic Visual Localization'' in \textit{IEEE/CVF Conference on Computer Vision and Pattern Recognition}, pp. 6896-6906, 2018.

\bibitem{ScanNet} A. Dai, A. X. Chang, M. Savva, M. Halber, T. Funkhouser and M. Nießner, ``ScanNet: Richly-Annotated 3D Reconstructions of Indoor Scenes,'' in \textit{IEEE Conference on Computer Vision and Pattern Recognition}, pp. 2432-2443, 2017.

\bibitem{Sizikova2016ECCVW} E. Sizikova, V. K. Singh, B. Georgescu, M. Halber, K. Ma and T. Chen, ``Enhancing Place Recognition Using Joint Intensity - Depth Analysis and Synthetic Data,'' in \textit{European Conference on Computer Vision Workshops}, 2016.

\bibitem{FDSLAM} X. Yang, Y. Ming and A. Calway, ``FD-SLAM: 3-D Reconstruction Using Features and Dense Matching,'' in \textit{IEEE International Conference on Robotics and Automation}, 2022.



\bibitem{LCD} F. Taubner, F. Tschopp, T. Novkovic, R. Siegwart and F. Furrer, ``LCD – Line Clustering and Description for Place Recognition,'' in \textit{International Conference on 3D Vision}, pp. 908-917, 2020.

\bibitem{SpoxelNet} M. Y. Chang, S. Yeon, S. Ryu and D. Lee, ``SpoxelNet: Spherical Voxel-based Deep Place Recognition for 3D Point Clouds of Crowded Indoor Spaces,'' in \textit{IEEE/RSJ International Conference on Intelligent Robots and Systems}, pp. 8564-8570, 2020.

\bibitem{Frampton2013BMVC} R. Frampton and A. Calway. ``Place Recognition from Disparate Views,'' in \textit{British Machine Vision Conference}, 2013.

\bibitem{Finman2014ICRA} R. Finman, T. Whelan, L. Paull and J. J. Leonard, ``Physical Words for Place Recognition in Dense RGB-D Maps,'' in \textit{IEEE International Conference on Robotics and Automation Workshop}, 2014.



\bibitem{Ming2021IROS} Y. Ming, X. Yang and A. Calway, ``Object-Augmented RGB-D SLAM for Wide-Disparity Relocalisation,'' in \textit{IEEE/RSJ International Conference on Intelligent Robots and Systems}, pp. 2180-2186, 2021.



\bibitem{Budvytis2019BMVC} I. Budvytis, P. Sauer and R. Cipolla, ``Semantic Localisation via Globally Unique Instance Segmentation,'' in \textit{British Machine Vision Conference}, 2019.

\bibitem{Garg2019JRR} S. Garg, N. Suenderhauf and M. Milford, ``Semantic-geometric Visual Place Recognition: A New Perspective for Reconciling Opposing Views,'' in \textit{International Journal of Robotics Research}, 2019.

\bibitem{Neubert2021RSS} P. Neubert, S. Schubert, K. Schlegel and P. Protzel, ``Vector Semantic Representations as Descriptors for Visual Place Recognition,'' in \textit{Robotics: Science and Systems}, 2021.

\bibitem{VLASE} X. Yu, S. Chaturvedi, C. Feng, Y. Taguchi, T. K. Lee, C. Fernandes and S. Ramalingam, ``VLASE: Vehicle Localization by Aggregating Semantic Edges,'' in \textit{IEEE/RSJ International Conference on Intelligent Robots and Systems}, pp. 3196-3203, 2018.


\bibitem{Orabona2007BMVC} F. Orabona and C. Castellini, ``Indoor Place Recognition Using Online Independent Support Vector Machine,'' in \textit{British Machine Vision Conference}, 2007.

\bibitem{Pronobis2010RAS} A. Pronobis, B. Caputo, P. Jensfelt and H.I. Christensen, ``A Realistic Benchmark for Visual Indoor Place Recognition,'' in \textit{Robotics and Autonomous Systems}, vol. 58, no. 1, pp. 81-96, 2010.


\bibitem{Song2019TIP} X. Song, S. Jiang, L. Herranz and C. Chen, ``Learning Effective RGB-D Representations for Scene Recognition,'' in \textit{IEEE Transactions on Image Processing}, vol. 28, no. 2, pp. 980–993, 2019.

\bibitem{Xiong2021TIP} Z. Xiong, Y. Yuan and Q. Wang, ``ASK: Adaptively Selecting Key Local Features for RGB-D Scene Recognition,'' in \textit{IEEE Transactions on Image Processing}, vol. 30, pp. 2722-2733, 2021.

\bibitem{Du2019CVPR} D. Du, L. Wang, H. Wang, K. Zhao and G. Wu, ``Translate-to-Recognize Networks for RGB-D Scene Recognition," in \textit{IEEE/CVF Conference on Computer Vision and Pattern Recognition}, pp. 11828-11837, 2019.


\bibitem{Huang2020IROS} S. Huang, M. Usvyatsov and K. Schindler, ``Indoor Scene Recognition in 3D,'' in \textit{IEEE/RSJ International Conference on Intelligent Robots and Systems}, pp. 8041-8048, 2020.

\bibitem{KPConv} H. Thomas, C. R. Qi, J. Deschaud, B. Marcotegui, F. Goulette and L. Guibas, ``KPConv: Flexible and Deformable Convolution for Point Clouds,'' in \textit{IEEE/CVF International Conference on Computer Vision}, pp. 6410-6419, 2019.

\bibitem{SIFT} D. G. Lowe, ``Distinctive Image Features from Scale-Invariant Keypoints,'' in \textit{International Journal of Computer Vision}, vol 60, no. 2, pp. 91-110, 2004.

\bibitem{BoW} J. Sivic and A. Zisserman, ``Efficient Visual Search of Videos Cast as Text Retrieval,'' in \textit{IEEE Transactions on Pattern Analysis and Machine Intelligence}, vol. 31, no. 4, pp. 591-606, 2009.




\bibitem{PointNet} R. Q. Charles, H. Su, M. Kaichun and L. J. Guibas, ``PointNet: Deep Learning on Point Sets for 3D Classification and Segmentation,'' in \textit{IEEE Conference on Computer Vision and Pattern Recognition}, pp. 77-85, 2017.



\end{thebibliography}
\end{document}